\pdfoutput=1
\documentclass[a4paper,11pt]{report}
\usepackage{float,graphicx}
\usepackage{amsfonts}
\usepackage{amsthm}
\usepackage{amsmath}
\usepackage{amssymb}
\usepackage{amstext}
\usepackage{wrapfig}
\usepackage{titling}
\usepackage[frozencache]{minted}
\usepackage{mathrsfs}
\usepackage[inline]{enumitem}
\usepackage{fancyvrb}

\usepackage[numbers,square]{natbib}

\usepackage{caption}
\usepackage{pgfplots}
\usepackage{mdframed}
\usepackage{multirow}
\usepackage[lined,boxruled]{algorithm2e}\usepackage{color}
\usepackage[english]{babel}

\newcommand*{\LongestText}{fermentum fringilla mauris }%
\newlength{\LargestSize}%
\usepackage[normalem]{ulem}

\pgfplotsset{width=10cm,compat=1.9}
\BeforeBeginEnvironment{minted}{\begin{mdframed}}
\AfterEndEnvironment{minted}{\end{mdframed}}

\title{\Large{\textbf{Inducing game rules from varying quality game play}}}
\date{}
\author{Alastair Flynn}

\begin{document}
\begin{titlingpage} 
\begin{center}

\begin{LARGE} 
\textbf{Inducing game rules from varying quality game play} \\
\end{LARGE}
\vspace{2cm}
\includegraphics[height=4cm]{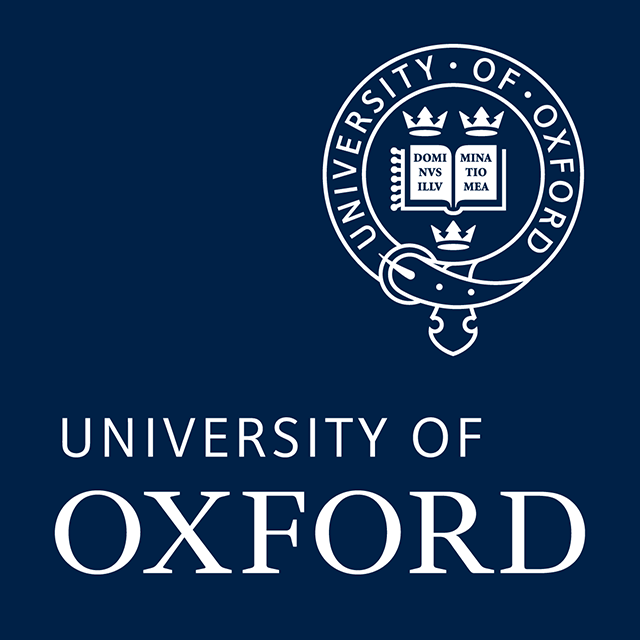}\\ 
\vspace{2cm}
Alastair Flynn\\
Hilary term, 2020
\end{center}
\end{titlingpage}

\tableofcontents

\begin{abstract}

General Game Playing (GGP) is a framework in which an artificial intelligence program is required to play a variety of games successfully. It acts as a test bed for AI and motivator of research. The AI is given a random game description at runtime (such as \textit{checkers} or \textit{tic tac toe}) which it then plays.
The framework includes repositories of game rules written in a logic programming language.

The Inductive General Game Playing (IGGP) problem challenges machine learning systems to learn these GGP game rules by watching the game being played. In other words, IGGP is the problem of inducing general game rules from specific game observations. Inductive Logic Programming (ILP), a subfield of ML, has shown to be a promising approach to this problem though it has been demonstrated that it is still a hard problem for ILP systems.

Existing work on IGGP has always assumed that the game player being observed makes random moves. This is not representative of how a human learns to play a game, to learn to play chess we watch someone who is playing to win. With random gameplay situations that would normally be encountered when humans play are not present. Some games rules may not come into effect unless the game gets to a certain state such as \textit{castling} in \textit{checkers}.

To address this limitation, we analyse the effect of using intelligent versus random gameplay traces as well as the effect of varying the number of traces in the training set.

We use Sancho, the 2014 GGP competition winner, to generate intelligent game traces for a large number of games. We then use the ILP systems, Metagol, Aleph and ILASP to induce game rules from the traces.
We train and test the systems on combinations of intelligent and random data including a mixture of both. We also vary the volume of training data.

Our results show that whilst some games were learned more effectively in some of the experiments than others no overall trend was statistically significant.

The implications of this work are that varying the quality of training data as described in this paper has strong effects on the accuracy of the learned game rules; however one solution does not work for all games.


\end{abstract}
\chapter{Introduction}\label{ch:intro}
General Game Playing (GGP) is a framework in which artificial intelligence programs are required to play a large number of games successfully \cite{Genesereth/GGPOverview}.
Traditionally game playing AI have focused on a single game \cite{AlphaGo,DeepBlue,Schaeffer/Checkers,Tesauro/Backgammon}. Famous AI include programs such as IBMs Deep Blue which is able to beat grand masters at chess but is completely unable to play checkers \cite{DeepBlue}. These traditional AI also only do part of the work. A lot of the analysis of the game is often done outside of the system \cite{Schaeffer/Checkers}. A more interesting challenge is building AI that can play games without any prior knowledge. In GGP the AI are given the description of the rules of a game at runtime. Games in the framework range greatly in both number of players and complexity; from the single player \textit{eight puzzle} to the six player \textit{chinese checkers}, and from the relatively simple \textit{rock paper scissors} to \textit{chess}  \cite{GGP-Website}. The progress in the field is consolidated annually at the GGP competition where participants compete to find the best GGP AI \cite{Genesereth/GGPOverview}.

The GGP framework includes a large database of games. In a GGP match games from these databases are selected at random and sent to the competitors \cite{Genesereth/GGPOverview}. The games are specified in the Game Description Language (GDL), a logic programming language built for describing games as state machines \cite{GDL_Spec}. An example of GDL rules is given in listing \ref{lst:GDL}.

\begin{figure}[ht]
	\centering
	\fbox{\includegraphics[width=0.7\linewidth]{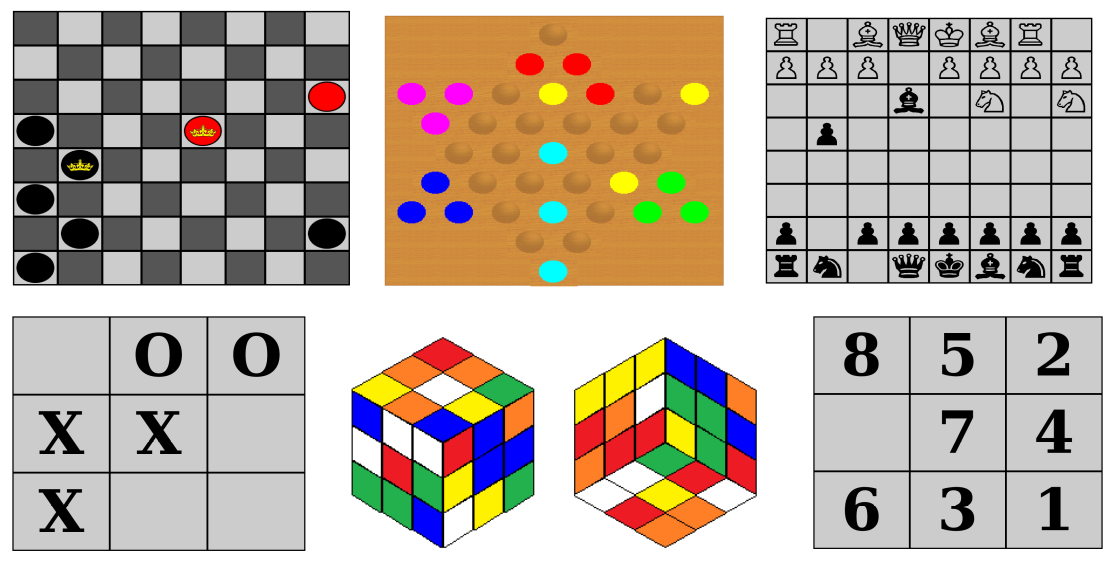}}
	\caption{A selection of games from the GGP competition. From the top left: \textit{checkers}, \textit{chinese checkers}, \textit{chess}, \textit{tic tac toe}, \textit{rubik's cube} and \textit{eight puzzle}}
\end{figure}

\begin{listing}[ht]
\begin{Verbatim}[frame=single,fontsize=\footnotesize]
(succ 0  1)
(succ 1  2)
(succ 2  3)
(beats scissors paper)
(beats paper stone)
(beats stone scissors)
(<= (legal ?p scissors) (player ?p))
(<= (legal ?p paper) (player ?p))
(<= (legal ?p stone) (player ?p))
(<= (draws ?p) (does ?p ?a) (does ?q ?a) (distinct ?p ?q))
(<= (wins ?p) (does ?p ?a1) (does ?q ?a2) (distinct ?p ?q) (beats ?a1 ?a2))
(<= (loses ?p) (does ?p ?a1) (does ?q ?a2) (distinct ?p ?q) (beats ?a2 ?a1))
\end{Verbatim}
\caption{
A sample of rules from the GDL description of Rock Paper Scissors. The $?$ indicates a variable and $<=$ indicates an implication with the first expression after it being the head and the conjugation of the rest making up the body
}
\label{lst:GDL}
\end{listing}

These GDL game descriptions form the basis for the Inductive General Game Playing (IGGP) problem \cite{Cropper/IGGP}. The task is an inversion of the GGP problem. Rather than taking game rules and using them to play the game in IGGP the aim is to learn the rules from observations of gameplay, similar to how a human might work out the rules of a game by watching someone play it. Cropper et al. \cite{Cropper/IGGP} define the IGGP problem in their 2019 paper; given a set of gameplay observations the goal is to induce the rules of the game. The games used in IGGP are selected from the GGP competition problem set meaning they are varied in complexity. An example IGGP problem for \textit{rock paper scissors} is given in table \ref{tab:IGGPTask} and the rules for the GGP game description in table \ref{tab:GGP-RPS}.

\begin{table}[]
	\centering
	\begin{tabular}{l|l|l}
		\multicolumn{1}{c|}{$BK$}                          & \multicolumn{1}{c|}{$E^+$}                                                                                      & $E^-$                                                                                                                                                                                                                                       \\ \hline
		\texttt{\begin{tabular}[c]{@{}l@{}}beats(scissors,paper).\\ beats(paper,stone).\\ beats(stone.scissors).\\ \\ succ(0,1).\\ succ(1,2).\\ succ(2,3).\\ \\ player(p1).\\ player(p2).\\ \\ true\_step(0).\\ true\_score(p1,0).\\ true\_score(p2,0).\end{tabular}} & \texttt{\begin{tabular}[c]{@{}l@{}}next\_score(p1,0).\\next\_score(p2,1).\\ \\ next\_step(1).\end{tabular}} & \texttt{\begin{tabular}[c]{@{}l@{}}next\_score(p1,1).\\ next\_score(p1,2).\\ next\_score(p1,3).\\ next\_score(p2,0).\\ next\_score(p2\_2).\\ next\_score(p2,3).\\ \\ next\_step(0).\\ next\_step(2).\\ next\_step(3).\end{tabular}}
	\end{tabular}
\caption{An example of an IGGP task for \textit{Rock Paper Scissors}. The task consists of the background knowledge ($BK$), the positive examples ($E^+$) and the negative examples ($E^-$). Here the predicate to be learned is \texttt{next}. We treat the \texttt{next score} and \texttt{next step} as two separate tasks}
\label{tab:IGGPTask}
\end{table}

\begin{table}[]
	\centering
	\begin{tabular}{|ll|}
		\hline
		\texttt{\begin{tabular}[c]{@{}l@{}}next\_step(N):-\\ \ \ true\_step(M),\\ \ \ succ(M,N).\\ next\_score(P,N):-\\ \ \ true\_score(P,N),\\ \ \ draws(P).\\ next\_score(P,N):-\\ \ \ true\_score(P,N),\\ \ \ loses(P).\\ next\_score(P,N2):-\\ \ \ true\_score(P,N1),\\ \ \ succ(N2,N1),\\ \ \ wins(P).\end{tabular}} & \texttt{\begin{tabular}[c]{@{}l@{}}draws(P):-\\ \ \ does(P,A),\\ \ \ does(Q,A),\\ \ \ distinct(P,Q).\\ loses(P):-\\ \ \ does(P,A1),\\ \ \ does(Q,A2),\\ \ \ distinct(P,Q),\\ \ \ beats(A2,A1).\\ wins(P):-\\ \ \ does(P,A1),\\ \ \ does(Q,A2),\\ \ \ distinct(P,Q),\\ \ \ beats(A1,A2).\end{tabular}} \\ \hline
	\end{tabular}
	\caption{The GGP rules for the \texttt{next step} and \texttt{next score} predicates. Translated from GDL to Prolog for readability.}
	\label{tab:GGP-RPS}
\end{table}

An effective way of solving the IGGP problem is a form of machine learning: Inductive Logic Programming (ILP)  \cite{Cropper/IGGP,Muggleton/ILP}. In ILP, the learner is tasked with learning logic programs given some background knowledge and a set of values for which the programs are true or false. In the IGGP paper, the authors showed through empirical evaluations that ILP systems achieve the best score in this task compared to other machine learning techniques \cite{Cropper/IGGP}. The ILP system derives a hypothesis, a logic program that when combined with the background knowledge entails all of the positive and none of the negative examples \cite{Muggleton/ILP}. In the IGGP paper it is also shown that the problem is hard for current ILP systems, with on average only 40\% of the rules being learned by the best performing systems \cite{Cropper/IGGP}.

However, the existing work has limitations. All work so far has assumed that the gameplay being observed is randomly generated \cite{Cropper/IGGP}. Rather than agents playing to win they simply make moves at random. Often this has the result of the game terminating before it reaches a goal state due to a cap on trace length or sections of the rule set remaining completely unused. There has also not been any research done to ascertain the effects that increasing the number of game observations has on the quality of the induced rules. It is unknown whether there is a threshold at which the adding new game traces does not further increase the accuracy. In this paper we evaluate the ability of ILP agents to correctly induce the rules of a game given different sets of gameplay observations - intelligent gameplay, random gameplay and combinations of the two. To generate the intelligent gameplay we use the system Sancho, winner of the 2014 GGP competition\footnote{http://ggp.stanford.edu/iggpc/winners.php accessed 28/04/2020} \cite{Sancho/Github}. 

Ideally in this paper we would compared optimal play versus random play or possibly human play against random play. However, it is clear gathering gameplay data at a consistent level human of play across a large set of game is not feasible. Gathering optimal gameplay data is also not realistic despite the fact that all the games have an optimal solution. The games used in this paper are finite, discrete, deterministic games of complete information, that is, games to which the minimax procedure can be applied. Using the minimax procedure to generate game traces would guarantee a form of optimality \cite{IntroToAlgorithms}. Namely each player would maximize their minimum gain. Despite this being an attractive prospect it is computationally infeasible to run this for the majority of games in the dataset. Checkers for example has a game tree far too large to reasonably compute \cite{Schaeffer/Checkers}. We could instead apply A* search, which is guaranteed to find an optimal solution when given an admissible heuristic. Instead Sancho was used since in the cases where the minimax procedure was easily solvable or known admissible heuristic were available for a game Sancho was able to provide optimal or near optimal play. It also brought the advantage of being able to play all the games indiscriminately.

Every year the GGP competition includes a ``Carbon versus Silicone'' event where humans pit themselves against the winner of the main competition. The machines almost always win \cite{Genesereth/GGPOverview}, so we can assume Sancho has above amateur human level play hence we use the word `intelligent' to describe its level of play. In this paper we also vary the number of gameplay traces from which the ILP systems learn to determine the optimal number of traces.

It is not obvious which of random or intelligent gameplay will achieve the best results. When learning the rules of chess would a human rather watch moves being made randomly, or a match between two grandmasters? It is not an easy question to answer. Both situations will result in a restricted view of the game, with certain situations never occurring in each one. This is not only a dilemma in the context of learning game rules. For example, teaching a self driving car to navigate roads requires training it on examples of driving. We would clearly not train it on highly intelligent Formula One quality driving and neither would we train it on random movement of the car. The question to be asked is what is the ideal quality level of training data to use to best teach a system the rules you want it to learn. In this paper, we try to help give some insight into this fundamental question. Specifically, we ask the following research questions:

\begin{description}
\item[Q1] Does varying the quality of game traces influence the ability for learners to solve the IGGP problems? Specifically, does the quality of game play affect predictive accuracy?
\item[Q2] Does varying the amount of game traces influence the ability for learners to solve the IGGP problems? Specifically, does the quality of game play affect predictive accuracy?
\item[Q3] Can we improve the performance of a learner by mixing the quality of traces?
\end{description}

We will train a range of ILP systems that each use a different approach to the problem on different sets of training data. The results for Q1 are the most interesting as it is not clear what the expected outcome is. Q2 has more of a natural answer. It is generally accepted that for machine learning problems the more training data you have the better the predictive accuracy of the ML system \cite{Mitchell/MachineLearing}. Often in ILP only a small amount of training data is needed, adding more data may not significantly affect accuracy \cite{Muggleton/ILP}.

The third question is interesting. Intuitively greater diversity in the training data should give a result closer to the rule that generated the data. However if a learner is trained on random data and only tested on random data we would expect this to perform better than a learner trained on a mix of intelligent and random data and then tested on random data since in the first case the training and test distributions match \cite{Mitchell/MachineLearing}. This question thus highlights an issue we face: how do we test the learned game rules?

Ideally the generated rules would be compared directly against the rules in the GDL game descriptions. We would take the generated rule and see for what percentage of all possible game states the reference rule and the learned rule gave the same output. Unfortunately we do not have the computational resources to do this with a lot of the games having too many possible states such as checkers which has a state-space complexity of roughly $5.0 \cdot 10^{20}$ \cite{Horssen/Checkers} and sudoku which exceeds $6.6 \cdot 10^{21}$ \cite{Felgenhauer/Suduko}. Instead we will test the learned programs on both intelligently generated and randomly generated data of the same quality used in training.

We would expect models trained on the same distribution as they are tested on to perform best since it is generally accepted that the accuracy of a model increases the closer the test data distribution is to the training data distribution  \cite{Mitchell/MachineLearing}. However, Gonzales and Abu-Mostafa \cite{Gonzalez/MismatchedOutperform} suggest that a system trained on a different domain to the one it is tested can outperform a system trained and tested on the same domain. Given a test distribution there exists a dual distribution that, when used to train, gives better results. The dual distribution gives a lower out-of-sample\footnote{out-of-sample data is data that is not in the training set} error than using the test distribution. This dual distribution can be thought of as the point in the input space where the least out-of-sample error occurs \cite{Gonzalez/MismatchedOutperform}.

As well as optimising the single training distribution we can take data from multiple distinct distributions. Ben-David et. al. \cite{Ben-David/DifferentDomains} show that training data taken from multiple different domains can in fact give lower error on testing data that training data taken from any single domain, including the testing domain. It is not clear in our case what selection of training data will result in the most effective learning.

To help answer questions 1-3, we make the following contributions:

\subsubsection{Contributions}
\begin{itemize}
\item We implement a system to play GGP games at random and highly intelligent levels (Chapter \ref{ch:traces}).
\item We transform the GGP traces to IGGP problems (Chapter \ref{ch:traces}).
\item We train the ILP systems Metagol \cite{Metagol/Github}, Aleph \cite{Muggleton/Aleph} and ILASP \cite{MarkLaw/ILASP2i, MarkLaw/OG-ILASP,MarkLaw/Thesis} on different combinations of intelligent and random data as well as testing them on each individually (Chapter \ref{ch:methodology}) to show that the difference in accuracy of the learned programs is small (Chapter \ref{ch:results}).
\item We train the ILP systems on differing quantities of game traces (Chapter \ref{ch:methodology}) to show through testing that an increase in the number of traces does not necessarily cause an increase in the accuracy of the learned programs (Chapter \ref{ch:results}).

\end{itemize}
\chapter{Related work}

\section{General Game Playing}

General Game Playing (GGP) is a framework for evaluating an agent's general intelligence across a wide range of tasks \cite{Cropper/IGGP,Genesereth/GGPOverview}. The agents accept declarative descriptions of arbitrary games at run time and use the descriptions to play those games effectively. All the games are finite, discrete, deterministic multi-player games of complete information \cite{GDL_Spec}. The games in the GGP competition game set vary in number of players, dimensions and complexity. For example games such as \textit{rock paper scissors} have 0 dimensions and only 10 rules in the given GGP rule set, more complex games such as \textit{checkers} has 52 rules and are 2 dimensional. There are also single player games such as \textit{eight puzzle} or \textit{fizz buzz}.

In 2005 an annual International General Game Playing Competition (IGGPC) was set up \cite{Kowalski/GGP}. Each year hopeful participants pit GGP agents against one another to determine the most effective system. The competitors take part in a series of rounds of increasing complexity. The agent that wins the most games in these rounds is declared victorious. The 2014 winner Sancho is used in this paper to generate traces of intelligent game play for the IGGP task.

The game are specified in the Game Description Language (GDL). These descriptions are used as example rule sets in the Inductive General Game Playing (IGGP) problem as the output that the learners would ideally generate. The descriptions of the games used in GGP are not necessarily minimal so it is possible that an ILP system could generate a more concise rule set than the GGP descriptions.

\subsection{Game Description Language}\label{sec:GDL}
Game Description Language (GDL) is the formal language used in the GGP competition to specify the rules of the games \cite{GDL_Spec}. The language is based on a logical programming language \textit{Datalog}, a subset of Prolog. To understand the semantics of GDL it helps to first cover logic programming as a field of study.
\subsubsection{Logic Programming}
Logic programming is a programming paradigm based on formal logic. Programs are made up of facts and rules. Rules are made up of two parts: the head and the body. They can be read as logical implications where the conjunction of all the elements in the body imply the head. The syntax is different for different logical programming languages but the head is usually written before the body in reverse implication. For example \texttt{a :- b,c}. A fact is simply a rule without a body, that is, a statement that is taken as true. The language interpreter takes logical statements as queries and returns whether they are true or false. If there are free variables in the query the interpreter assigns them values for which the query is true. Logical programming is good for symbolic non-numeric computation \cite{Bratko}. It is well suited to solving problems that involve well defined objects and relations between them, such as a GGP game.
\subsubsection{Usefulness of GDL}
The game description language was designed specifically to represent finite, discrete, deterministic multi-player games of complete information. The language is based on \textit{Datalog} and the property of any question of logical entailment being decidable is retained.

The games are defined in terms of a set of true facts and logical rules that capture the information needed to give the following predicates:
\begin{itemize}
\item The initial game state
\item The goal states
\item The terminal states
\item The legal moves for a given player from a given state
\item The next state for a given player, state and move
\end{itemize}

The GDL language is an extension of \textit{Datalog$^{\neg}$}, that is \textit{Datalog} with stratified negation \cite{GDL_Spec}. \textit{Datalog} allows only universally quantified rules consisting of a conjunction of positive atoms that imply a single atom. \textit{Datalog$^{\neg}$} allows for negative as well as positive atoms \cite{Alice/Foundations}. GDL also allows for some functional symbols, that is predicates containing other predicates however it restricts to keep the finite model property that is inherent to \textit{Datalog} \cite{GDL_Spec}.

In GDL variables are written with the symbol \texttt{?} before them and rules are written starting with \verb|=>| followed by the head followed by the body predicates which are interpreted as a conjunction. All rules are universally quantified. For example
\[\texttt{(<= (next (cell ?x ?y ?player)) (does ?player (mark ?x ?y)))}\]
in first order logic this would be written
\[\forall x. \forall y. \forall player. does(player,(mark(x,y))) \rightarrow cell(x,y,player)\]
GDL also reserves certain words as listed in table \ref{tab:GDL}
\begin{center}
	\begin{table}
	\begin{tabular}{| l | l |}
		\hline
		\texttt{true(?f)} & Atom \texttt{?f} is true in the current game state \\ \hline
		\texttt{does(?r,?m)} & Player \texttt{?r} performs action \texttt{?m} in the current state \\ \hline
		\texttt{next(?f)} & Atom \texttt{?f} will be true in the next game state \\ \hline
		\texttt{legal(?r,?m)} & Action \texttt{?m} is a legal move for player \texttt{?r} in the current game state\\ \hline
		\texttt{goal(?r,?n)} & Player \texttt{?r} performs action \texttt{?m} in the current game state\\ \hline
		\texttt{terminal} & The current state is terminal\\ \hline
		\texttt{init(?f)} & Atom \texttt{?f} is true in the initial game state\\ \hline
		\texttt{role(?n)} & The Constant \texttt{?n} denotes a player\\ \hline
		\texttt{distinct(?x,?y)} & \texttt{?x} and \texttt{?y} are syntactically different\\
		\hline

	\end{tabular}
	\caption{The reserved predicates in GDL}
\label{tab:GDL}
\end{table}

\end{center}

In the IGGP problem (section \ref{sec:IGGP}) the given task is to generate the rules to predict the goal, next, legal and terminal predicates.

\section{Inductive General Game Playing}\label{sec:IGGP}
The Inductive General Game Playing (IGGP) problem is an inversion of the GGP problem. Rather than using game rules to generate gameplay the learner must learn the rules of the game from observations of game play. The learner is given a set of game traces and is tasked with using them to induce (learn) the rules of the game that produced the traces \cite{Cropper/IGGP}. IGGP was designed as a way of benchmarking machine learning systems.

The definition of task itself is based on the Inductive Logic Programming problem.

\subsection{Inductive Logic Programming}\label{sec:ILP}
Inductive Logic Programming (ILP) is a form of machine learning that uses logic programming to represent examples, background knowledge, and learned programs \cite{Cropper/EfficientLearning}. To learn the ML system is supplied with positive examples, negative examples and the background knowledge. In the general inductive setting we are provided with three languages.
\begin{itemize}
\item $\mathcal{L}_O$: the language of observations (positive and negative examples)
\item $\mathcal{L}_B$: the language of background knowledge
\item $\mathcal{L}_H$: the language of hypotheses
\end{itemize}
The general inductive problem is as follows: given a set of positive examples $E^+ \subseteq \mathcal{L}_O$, negative examples $E^- \subseteq \mathcal{L}_O$ and  background knowledge $B \subseteq \mathcal{L}_B$ find an hypothesis $H \in \mathcal{L}_H$ such that 
\[B \cup H \models E^+ \text{ and } B \cup H \not\models E^-\]
\cite{Muggleton/ILP}
That is that the generated hypothesis and the background knowledge imply the positive examples and do not imply the negative examples.

\subsection{Back to IGGP}

In IGGP we also have the idea of background knowledge and positive or negative observations. The task itself is closely based on the ILP problem and is described in chapter \ref{ch:IGGP}. The games used for the IGGP problem are taken from the IGGP dataset. It is a collection of 50 games, specified in GDL. The purpose of this database is to standardise the set of games used in the IGGP problem to allow for results to be easily compared. It is the set of games used in this paper.
A mechanism is also provided by the authors to turn these GDL game descriptions in the set into new IGGP tasks. This mechanism simulates random play of the games to generate the observations. In this paper we modify it to generate optimal game traces.

\section{ILP systems used}
We use three ILP systems to compare the effects of different learning data, Metagol, Aleph and ILASP. There are many approaches to the ILP problem \cite{Svetla/ILPOverview,Cropper/NewIdeas}. The three systems here all represent different approaches to the problem but certainly do not give a full representation of the techniques available.
\subsection{Metagol}
The Metagol ILP system is a meta-interpreter for Prolog, that is, it is written in the same language is evaluates \cite{Cropper/Thesis,Rolf/Metagol,Metagol/Github}. Metagol takes positive and negative examples, background knowledge and metarules. Metarules are specific to Metagol. They determine the shape of the induced rules and are used to guide the search for a hypothesis. An example of a metarule is the \textit{chain} rule \[P(A,B) \leftarrow Q(A,C),R(C,B)\] The letters $P$, $Q$ and $R$ represent existentially quantified second order variables, $A$, $B$ and $C$ are regular first order variables. When trying to induce rules the second order variables are substituted for predicates from the background knowledge or the hypothesis itself. To illustrate this consider a metarule being applied when learning the predicate \textit{last(A,B)} where a is a list and b is the last element in it. Given the positive example
\begin{minted}{Prolog}
last([a,l,g,o,r,i,t,h,m],m).
\end{minted}

As well as the background predicates \textit{reverse/2} and \textit{head/2} the chain rule might be used to derive the rule
\begin{minted}{Prolog}
last(A,B) :- reverse(A,C), head(C,B)
\end{minted}

\begin{enumerate}
\item Select a positive example (an atom) to generalise. If none exists, stop, otherwise proceed to the next step.
\item Try to prove an atom using background knowledge by delegating the proof to Prolog. If successful, go to step 1, otherwise proceed to the next step.
\item Try to unify the atom with the head of a metarule and either choose predicates from the background knowledge that imply the head to fill the body. Try proving the body predicate, if it cannot be proved try different predicate in the background knowledge\footnote{To prove the body predicate the whole procedure is called again with the body predicates as the positive examples. For example if we had \texttt{last([a,b],b)} as our positive example and have tried to use the chain metarule to to prove it with reverse and head we would then call the whole procedure again with the positive examples as \texttt{[reverse([a,b],C),head(C,b)]} if this was successful then we continue, otherwise we try different predicates}. If no predicate can be found that proves the positive example then try adding a new invented predicate and attempt to prove this\footnote{For example we might replace head with an invented predicate in the previous footnote example}.
\item Once you find a metarule substitution that works add it to the program and move to the next atom
\end{enumerate}

The end hypothesis is all the metarule substitutions. It is then checked that the negative atoms are not implied by the hypothesis, if they are a new one is generated. When the hypothesis is combined with the background knowledge the positive examples, but not the negative examples, are implied.

The choice of metarules determines the structure of the hypothesis. Different choices of metarules will allow different hypotheses to be generated. Deciding which metarules to use for a given task is an unsolved problem  \cite{Cropper/Thesis}. For this task a set of 9 derivationally irreducible metarules are used which remain consistent across all tasks.

\subsection{Aleph}\label{sec:aleph}

Aleph is a Prolog variant of the ILP system Progol  \cite{Muggleton/Aleph}. As input Aleph takes positive and negative examples represented as a set of facts along with the background knowledge. It also requires \textit{mode declarations} and \textit{determinations} which are specific to Aleph. Mode declarations specify the type of the inputs and outputs of each predicate used. Determinations specify which predicates can go in the body of a hypothesis. These determinations take the form of pairs of predicates, the first being the head of the clause and the second a predicate that can appear in its body.

For each predicate we would like to learn in the IGGP problem we give Aleph the determinations consisting of every target predicate (next, goal and legal) paired with every background predicate (which are specific to each game). Luckily there has been some work to induce mode declarations from the determinations  \cite{McCreath/Meta-extraction} so we do not need to come up with our own mode declarations.

A basic outline of the Aleph algorithm is taken from the aleph website \footnote{http://www.cs.ox.ac.uk/activities/programinduction/Aleph/aleph.html accessed 26/03/2020}:
\begin{enumerate}
\item Select an example to be generalised. If none exist, stop, otherwise proceed to the
next step.
\item Construct the most specific clause (also known as the bottom clause  \cite{Muggleton/Aleph}) that entails
the example selected and is within language restrictions provided.
\item Search for a clause more general than the bottom clause. This step is done by searching for some subset of the literals in the bottom clause that has the 'best' score.
\item The clause with the best score is added to the current theory and all the examples
made redundant are removed. Return to step 1.
\end{enumerate}

Mode declarations and determinations are used in step 2 of this procedure to bound the hypothesis space. Only predicates that are in the determinations of the hypothesis and are of the correct type are tried. The bottom clause constructed is the most specific clause that entails the example. Therefore a clause with the same head and any subset of the predicates of the body will be more general than the bottom clause. Aleph only considers these generalisations of this bottom clause.

By default Aleph preforms a bounded breath first search on all the possible rules, enumerating shorter clauses before longer clauses. The search is bounded by several parameters such as maximum clause size and maximum proof depth.

\subsection{ILASP}

ILASP is an ILP system based on Answer Set Programming (ASP). An introduction to ASP can be found here  \cite{Corapi/ASP}. ILASP uses a subset of ASP that is defined in these papers  \cite{ILASP-Manuel,MarkLaw/OG-ILASP,MarkLaw/Thesis} The ILASP process effectively generates all possible rules of a certain length, turns the problem into an ASP problem that adds a predicate to each rule allowing it to be active or inactive. It then uses the ASP solver Clingo \cite{Clingo}  to check which rules should be active if the program is to be consistent with the positive and the negation of the negative examples  \cite{MarkLaw/OG-ILASP,MarkLaw/Thesis}. In this paper we use a version of ILASP based on ILASP2i  \cite{MarkLaw/ILASP2i} which was developed with the IGGP problem in mind \cite{Cropper/IGGP}. As one input ILASP takes a hypothesis search space, i.e. the set of all hypotheses to be considered. This is constructed using the type signatures given for each problem that are provided in the IGGP dataset.

An ILASP task is defined as a tuple $T = \langle B,S_M,E^+E^-\rangle$ where $B$ is the background knowledge, $S_M$ is the hypothesis space, and $E^+$ and $E^-$ are the positive and negative examples. The ILASP procedure is given in algorithm \ref{alg:ILASP}.
\begin{algorithm}[H]\label{alg:ILASP}
    \SetAlgoLined
    $n$ = 0\;
    solutions = []\;
    \While{solutions.empty}{
        $S^N$ = all possible hypotheses of length $N$ from $S_M$ \;
        $ns$ = all subsets of $S^N$ that imply $E^-$ (Using an ASP solver)\;
        $vs$ = the set of rules that for each set of rules in $ns$ imply false if exactly those rules are active\;
        solutions = all subsets of $S^N$ that imply $E^+$ and satisfy $vs$ (using asp solver)\;
        $n$ = $n$ + 1\;
    }
    \caption{ILASP outline}
\end{algorithm}

The approaches used by ILASP have proved to be well suited to the IGGP task  \cite{Cropper/IGGP}.
\chapter{IGGP problem}
\label{ch:IGGP}

The IGGP problem is defined in the 2019 paper Inductive General Game Playing \cite{Cropper/IGGP}. Much like the problem of ILP described in section \ref{sec:ILP} the problem setting consists of examples about the truth or falsity of a predicate $F$ and a hypothesis $H$ which covers $F$ if $H$ entails $F$. We assume the languages of:
\begin{itemize}
\item $\mathscr{E}$ the language of examples (observations)
\item $\mathscr{B}$ the language of background knowledge
\item $\mathscr{H}$ the language of hypotheses
\end{itemize}
Each of these languages can be seen as a subset of those defined for the ILP task. In our experiments we transform data generated from the GDL descriptions of games in the IGGP dataset and transform it into the languages of $\mathscr{B}$ and $\mathscr{E}$. GDL allows for functional symbols in rules albeit in a restricted form. For example any atom appearing inside a \texttt{true} predicate such as \texttt{true(count(9))}. We flatten all of these to single, non nested predicates, i.e. \verb|true_count(9)|. This is needed as not all ILP systems support function symbols. We can therefore assume that both $\mathscr{E}$ and $\mathscr{B}$ are made up of function-free ground atoms. The language of hypotheses $\mathscr{H}$ can be assumed to consist of datalog programs with stratified negation as described here \cite{Kenneth}. Stratified negation is not necessary but in practice allows significantly more concise programs, and thus often makes the learning task computationally easier. We first define an IGGP \textit{input} then use it to define the IGGP \textit{problem}. The input is based on the general input for the Logical induction problem.

\textbf{The IGGP Input:} An input $\Delta$ is a set of triples $\{(B_i,E_i^+,E_i^-)\}_{i=0}^m$ where
\begin{itemize}
\item $B_i \subset \mathscr{B}$ represents background knowledge
\item $E_i^+ \subseteq \mathscr{E}$ and $E_i^- \subseteq \mathscr{E}$ represent positive and negative examples respectively
\end{itemize}
The IGGP input composes the IGGP problem as follows:

\textbf{The IGGP Problem:} Given an IGGP input $\Delta$, the IGGP problem is to return a hypothesis $H \in \mathscr{H}$ such that for all $(B_i,E_i^+,E_i^-) \in \Delta$ it holds that $H \cup B_i \models E_i^+$ and
$H \cup B_i \not\models E_i^-$.

In this paper we use the IGGP problem to analyse the ability of ILP agents to learn from a range of training data of differing quality. To analyse the ability of the ILP systems to learn we use, for each IGGP task $\Delta$ and hypothesis $H$, a testing set $T = T^+\cup T^-$ where
\begin{itemize}
	\item $T^+ \subseteq \mathscr{E}$ and $T^-\subseteq \mathscr{E}$ are the positive and negative testing observations.
	\item $T^+ \cap \bigcup_{i=0}^m E_i^+  = \emptyset$. The positive testing examples are distinct from the positive training examples
	\item $T^- \cap \bigcup_{i=0}^m E_i^-  = \emptyset$. The negative testing examples are distinct from the negative training examples
\end{itemize}
To test the systems we check for each $t^+ \in T^+$ whether\[ H \cup \bigcup_{i=0}^m B_i \models t^+\] and for each $t^- \in T^-$  whether \[ H \cup \bigcup_{i=0}^m B_i \not\models t^-\]
The success of the system is defined as the percentage of correctly classified $t \in T$. We use this definition of success to answer the research questions in chapter \ref{ch:intro}. The experiments performed to answer these questions are described in chapter \ref{ch:methodology}.

\chapter{Generating Traces}\label{ch:traces}
In this section we describe the process of generating game traces then transforming them into to IGGP tasks. The first step is to generate the intelligent and random gameplay. To do this we used the Sancho system \cite{Sancho/Github}.

\section{Sancho}\label{sec:sancho}
Generating intelligent gameplay is a nebulous task. Defining intelligence is no small ask. Using optimal gameplay traces or human generated gameplay would be ideal, however this is infeasible in the scope of this project. Provably optimal traces were experimented with. An optimal player for \textit{eight puzzle} using A* with an admissible heuristic was created with the intention of writing optimal players for more games such as \textit{tic tac toe} and \textit{knights tour}. However when testing the Sancho system it was found that game traces produced were often of near optimal quality. In fact for \textit{eight puzzle} it generates provably optimal traces as described later in the chapter. It was decided that it would be easier and we would generate a possibly more a consistent level of play across all games using the GGP player Sancho.

The winner of the 2014 GGP competition ``Sancho''\footnote{http://ggp.stanford.edu/iggpc/winners.php accessed on 12/03/2020} was selected since it is the most recent winner with published code \cite{Sancho/Github}. Some small modifications to the information logged by the game server are made but otherwise the code is unchanged. Since the aim of the GGP competition is to find the program that performs best at the set of games used in the IGGP problem we conclude that this system was the best general game player to use. We can assume Sancho can play above amateur human level across the majority of games in the the dataset. At the annual GGP competition there is also a match held at the end where the human creators of the AI themselves play against the winner. The humans almost never win \cite{Genesereth/GGPOverview}.

Sancho uses a range of techniques to generate gameplay. The core of algorithm used by Sancho is the Monte Carlo tree search (MCTS).

\subsection{Monte Carlo Tree Search}
Given a game state the basic MCTS will return the most promising next move. The algorithm achieves this by simulating random playouts of the game many times. The technique was developed for Computer Go but has since been applied to play a wide range of games effectively including board games and video games \cite{Silver/MCTS,Chaslot/MCTS}. The use of random simulation to evaluate game states is a powerful tool. No extra information about the game such as heuristics for evaluating states are needed at all, the rules of the game suffice. In the case of the GGP problem this is ideal.

In our case all games being played are sequential, finite and discrete so we only need to consider MCTS for this case.

\begin{figure}[ht]
	\centering
	\includegraphics[width=0.9\linewidth]{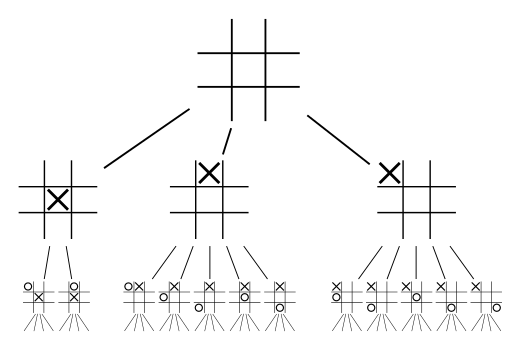}
	\caption{A section of a game tree of \textit{tic tac toe} showing some of the possible moves. \textit{Stannered/CC-BY-SA-3.0}}
	\label{fig:game-tree}
\end{figure}

The MCTS is a tree search algorithm, the tree being searched is the game tree. Each node on the game tree represents a state (figure \ref{fig:game-tree}). The search is a sequence of traversals of the game tree. A \textit{traversal} is a path that starts at the root node and continues down until it reaches a node that has at least one unvisited child (not necessarily a terminal state). One of these unvisited children is then chosen to be the start state for a simulation of the rest of the game. The simulation selects moves randomly, playing the game out to a terminal state. The result of the simulation is propagated back from the node it started at all the way to the root node, updating statistics attached to each node. These statistics are used to choose future paths to traverse so that more promising moves are investigated with higher probability.

\subsubsection{Use of MCTS in Sancho}

Sancho makes a few modifications to MCTS\footnote{https://sanchoggp.blogspot.com/2014/05/what-is-sancho.html accessed on 15/03/2020}. The main one being the adding of heuristics. It also makes optimisations to increase efficiency.

In GGP matches a period of time before the match is given in which to do pre match calculations. Sancho uses this period to derive basic heuristics. A heuristic should take a game state or move and return a value based on how promising that state or move is in relation to the goal state. 

The identification of possible heuristics takes place in two stages. The first of these is static analysis of the game rules. This static analysis identifies possible heuristics that can be applied to the current game. These include things like piece capture: if certain rules indicate capture of the players piece these can be selected against. Another is numeric quantity detection: a number in the state acts as a heuristic (like number of coins a player has). The second stage is simulation of the game. Many (possibly tens of thousands) of full simulations of the game are run. After the simulations are complete a correlation coefficient is calculated between each candidate heuristic's observed values and the eventual score achieved in the game. Those heuristics that show correlation above a fixed threshold are then enabled, and will be used during play to guide the tree exploration in MCTS.

\subsection{Other aspects of Sancho}

Monte Carlo simulations are used for most of the games available however Sancho identifies some games that can be solved more efficiently in other ways. Any single player game in the GGP dataset can be analysed using puzzle solving techniques.
\subsubsection{Puzzle Solving}
A single player game in the GGP dataset is necessarily a deterministic game of complete information due to the constraints of the Game Description Language. This gives the useful constraint that any solution found in a playout of the game will always remain valid since the game is completely deterministic. Sancho identifies these single player games and attempts to derive heuristics as described above. Where an obvious goal state exists a distance metric such as the hamming distance between two states can be applied. A* search is then used to find optimal solutions. For some games, such as \textit{eight puzzle} the hamming distance is an admissible heuristic so Sancho finds a provably optimal solution.

\subsubsection{Static analysis}

Sancho also does static analysis of the game rules. It determines game predicates that if true imply that either the goal will never be reached or the goal will always be reached. An example of this is a game such as \textit{untwisty corridor} (the game is effectively a maze where you immediately lose if you step off a ``safe'' path). If the safe path is stepped off then the goal can never be reached so any state not on the safe path is avoided.

\subsection{Sancho and IGGP}
Sancho generally provides good quality intelligent play when run on the games in the IGGP dataset however in some cases it struggles.

The dataset contains several games based on game theory. Games such as the \textit{prisoner dilemma} where player can choose to either \textit{cooperate} or \textit{defect}. If both players cooperate then they both get 3 points. If one defects and the other tried to cooperate the cooperator gets 0 points and the defector gets 5. If they both defect both get 0. There are several rounds of this in one game playout. A human playing this might choose to try and cooperate, hoping that that the other player will too however Sancho has been observed to always defect. This result is predictable since this is the only strong Nash equilibrium for this game however it is hard to justify this as ``human quality play''.

\section{Generating and transforming the traces}\label{sec:gen}

To generate the intelligent and random traces for the experiments an automatic system was constructed. The system runs Sancho or a random game player on the GDL game descriptions and converts the traces into IGGP tasks.

The general game playing community have developed a codebase that provides the basic functionality for hosting a match between GGP agents\footnote{https://github.com/ggp-org/ggp-base accessed on 13/03/2020}. This codebase along with the GGP agent Sancho is used to generate the datasets. The codebase includes a very basic random GGP player that will, given a GDL game description, play a random legal move every turn. To generate the random traces for a game of $x$ players $x$ Sancho or random GGP players were pitted against one another.

For the IGGP problem information about games is represented as sets of ground atoms. These atoms can be divided into two sets, the atoms that can change during the game and the atoms that can not. The set of atoms that can change can again be divided in two, the action atoms $A$ and the state atoms $S$.

The set $S$ represents all atoms that can change from one state to the other. Examples of elements of this set $S$ might be the state of the board in \textit{tic tac toe} along with which of the players is to take their turn next:
\begin{verbatim}
( control noughts )
( cell 1 1 b ) ( cell 1 2 x ) ( cell 1 3 o )
( cell 2 1 x ) ( cell 2 2 o ) ( cell 2 3 o )
( cell 3 1 b ) ( cell 3 2 x ) ( cell 3 3 b )
\end{verbatim}

The set $A$ is the set of ground atoms representing the actions that can be taken. The moves that each player makes are taken from this set. For a game such as \textit{eight puzzle} these would consist of the set of atoms:
\[\{\texttt{( move i j )}| 0<i\leq 9, 0<j\leq 9\}\]
Each atom represent sliding the tile at $(i,j)$ into the space on the board. To represent a player not making a move we also always have $noop \in A$.

\subsubsection{Game traces}

We modified the GGP matchmaker to log the following information:
\begin{itemize}
	\item \textbf{Game state trace} - The sequence of game states: $states = (s_1,...,s_n)$ where each $s_i \subseteq S$ is the set of ground atoms true in the $i^{th}$ state.
	\item \textbf{Game roles} - The list of roles of each player in the game: $roles = (r_1,...,r_k)$ e.g. \textit{noughts} and \textit{crosses} in \textit{tic tac toe}
	\item \textbf{Move trace} - The sequence of moves made by each player after each state: $moves = ((m_{1,r_1},...,m_{1,r_k}),...,(m_{n,r_1},...,m_{n,r_k}))$ where $m_{i,r_j} \in A$ is the $i^{th}$ move of player $r_j$. In games where not all players move every turn then $m_{i,r_j}=noop$ shows that $r_j$ made no move
	\item \textbf{Legal move trace} - The sequence of the legal moves for each player in each state:  $legal = ((l_{1,r_1},...,l_{1,r_k}),...,(l_{n,r_1},...,l_{n,r_k}))$. $l_{i,r_j} \subseteq A$ is the set of possible moves for $r_j$ in state $s_i$
	\item \textbf{Goal value trace} - The sequence of goal value for each player on every state: $goals = ((g_{1,r_1},...,g_{1,r_k}),...,(g_{n,r_1},...,g_{n,r_k}))$. $g_{i,r_j} \in [0,100]$ represents how well player $r_j$ has achieved the goal in state $s_i$.

\end{itemize}
This represents all the data needed from the match to generate IGGP tasks. In each state all the positive atoms in $S$ (that is the atoms true in the current state) are arguments of the \texttt{true} predicate. For example if at the start of the game the first cell is blank on the board then we would have $\texttt{( true ( cell 1 1 b ) )} \in s_1$. All the positive atoms in $A$ are arguments of the \texttt{does} predicate in a similar fashion.

\subsection{Transforming game traces into IGGP tasks}

We define a function that transforms the sequences into four separate induction tasks. An induction task is the set of triples $\{(B_i,E_i^+,E_i^-)\}_{i=0}^n$. We first define a function trace that translates the sequences given in the log of the match into a set of pairs $\{(B_i,E_i^+)\}_{i=0}^n$ . We then define a function \textit{triple} which gives the the set of triples  $\{(B_i,E_i^+,E_i^-)\}_{i=0}^n$ from the set of pairs $\{(B_i,E_i^+)\}_{i=0}^n$.
\\

We flatten the sequences of $(m_{i,r_1},...,m_{i,r_k})$ in \textit{moves}, $(l_{1,r_1},...,l_{1,r_k})$ in \textit{legals} and $(g_{1,r_1},...,g_{1,r_k})$ in \textit{goals} to a set that includes the role name as an extra argument of the predicate (the arity is increased by 1). For example if
\[(m_{i,r_1},m_{i,r_2}) = \texttt{[( move 2 2 ), ( move 2 3 )]}\ \ \ \ \ \ \ \ \ \ \ \ \ \ \ \ \ \ \ \ \ \ \]
and $r_1 = black$ and $r_2 = white$ then it would be replaced by $m_i$
\[m_i = \texttt{\{( move black 2 2 ), ( move white 2 3 )\}}\ \ \ \ \ \ \ \ \ \ \ \ \ \ \ \ \]
\textit{moves} is now a sequence $(m_1,...,m_n)$. With \textit{legals} and \textit{goals} a similar intuitive flattening procedure is applied. All four predicates are now represented by a single flat list.
\subsubsection{Building the traces}
The trace function is defined for \textit{legal}, \textit{goal}, \textit{next} and \textit{terminal}. We treat the output of the zip function, which takes a pair of lists and turns them into a list of pairs, as a set rather than a list. We define $S[p/q]$ on a set $S$ for predicates $p$ and $q$ to be the substitution of all instances of $q$ in $S$ for $p$.
\begin{align*}
&\Lambda_{legal} &&= trace_{legal}(states,legal) &&= \text{zip } states\ legal[\texttt{legal}/\texttt{true}] \\
&\Lambda_{goal} &&= trace_{goal}(states,goals) &&= \text{zip \textit{states goal}} \\
&\Lambda_{next} &&= trace_{next}(states,moves) &&= \text{zip \textit{states} } \text{(tail }states[\texttt{next}/\texttt{true}]) \\
&\Lambda_{terminal} &&= trace_{terminal}(states) &&= \text{zip \textit{states t}} \\
&&&\ \ \ \ \textbf{where  } t =\text{take } n\ (\text{repeat } \emptyset ) &&\text{++}\ \ [\{\texttt{( terminal )}\}] \\
&&&\ \ \ \ \textbf{and } n = (\text{length }states) - 1 &&
\end{align*}

The substitutions $[\texttt{legal}/\texttt{true}]$ and $[\texttt{next}/\texttt{true}]$ ensure that all positive examples in $E^+$ are in the relevant predicate for training. Before, the list \textit{legal} consisted of the only atoms in the predicate \texttt{true}. For the ILP systems to identify this as an example of legal we need to surround these with the \texttt{legal} predicate. That is, we need all the \texttt{move} atoms in $legal$ to be inside the \texttt{legal} predicate, e.g.:
\[\texttt{( legal ( move 1 1 ) )}\]
Since the predicates are already in the \texttt{true} predicate we only need to substitute one out for the other.

Some of the ILP systems being tested cannot work with function symbols of arity greater than 0. To allow them to operate we merge the function and their arguments into one predicate. For example \texttt{( legal ( move 1 1 ) )} becomes \verb|( legal_move 1 1 )| where \verb|legal_move| is a newly formed predicate.

To generate the triples we use the two functions $triples_1$ and $triples_2$. Let $pred\ X\ p$ be the subset of atoms in the set $X$ that use the predicate $p$.
\begin{align*}
tri&ple_1(\Lambda_p) = \text{map } f\ \Lambda\\
&\textbf{where } f\ (B,E^+) = (B,E^+,(pred\ S\ p) - E^+) \ \ \ \ \ \ \ \ \ \ \ \ \ \ \ \ \ \ \ \ \ \ \ \ \ \ \ \ \ \ \ \ \ \ \ \ \ \\
tri&ple_2(\Lambda_p) = \text{map } f\ \Lambda \\
&\textbf{where } f\ (B,E^+) = (B,E^+,(pred\ A\ p) - E^+)
\end{align*}
We generate the IGGP task with $triples_1$ and $triples_2$ as below:
\[\Delta = triples_1(\Lambda_{goal}) \cup triples_1(\Lambda_{terminal}) \cup triples_2(\Lambda_{next}) \cup triples_2(\Lambda_{legal})\]
The IGGP task $\Delta$ is set to the ILP systems as described in chapter \ref{ch:methodology}.

\chapter{Experimental methodology}\label{ch:methodology}

In this section we describe the exact methods used to run the experiments and the experiments themselves. Two experiments were run. Experiment E1 answers questions Q1 and Q3. The ILP systems are trained on random, intelligent and mixed quality traces. In experiment E2 they were trained on a varying number of mixed quality traces. 

\section{Materials}
\subsubsection{Generating training data}
The training data was generated according to the methods described in chapter \ref{ch:traces}. To generate the game traces the Sancho version 1.61 was used \cite{Sancho/Github}. The only modifications made to this were the changes to the logging system as described in section \ref{sec:gen}.

To generate traces matches between multiple random players or multiple instances of Sancho were conducted. Each match was run with 30 seconds pre game warm up time and a maximum of 15 seconds per move. The games used are listed in table \ref{tab:Games}. Due to issues of compatibility only 36 out of the 50 games from the IGGP dataset were used. Some games caused issues with Sanchos simulations and thus is was not possible to run experiments with them. 

All games used from the IGGP dataset have a limit on the number of moves however it varied from game to game.

Due to restraints on computational power 30 random traces and 30 Sancho generated traces were generated. The traces were generated on a machine with 20GB of RAM and an 8 core processor. 

\subsubsection{The ILP systems}

The three ILP systems were trained on the random, intelligent and mixed game traces using the same settings as were used in the IGGP paper \cite{Cropper/IGGP}. These settings are:
\begin{itemize}
	\item Metagol - Metagol 2.2.3 with YAP 6.2.2. The metarules used can be found in the IGGP code repository\footnote{https://github.com/andrewcropper/mlj19-iggp accessed 23/05/20}.
	\item Aleph - Aleph 5 with YAP 6.2.2. The default Aleph parameters were used.
	\item ILASP - A specialised version ILASP* developed for this task was used and can be found in the IGGP code repository. It is based on ILASP2i.
\end{itemize}
The systems were run concurrently in both training and testing.

\begin{table}[]
\begin{tabular}{|l|l|l|l|}
	\hline
	alquerque           & dont touch    & gt ultimatum              & pentago\\ \hline
	asylum              & duikoshi      & hex for   three               & platform jumpers\\ \hline
	battle of numbers   & eight puzzle  & horseshoe                 & rainbow\\ \hline
	breakthrough        & farming       & hunter                    & sheep and wolf\\ \hline
	buttons and lights  & fizz buzz      & knights tour              & sudoku\\ \hline
	centipede           & forager       & kono                      & sukoshi\\ \hline
	checkers            & gt centipede  & light board                & tic tac toe \\ \hline
	coins               & gt chicken    & multiple buttons and lights  & ttcc4 \\ \hline
	connect4team        & gt prisoner   & nine board tic tac toe        & untwisty corridor \\ \hline
\end{tabular}
\caption{Games used in the experiments. \texttt{gt} stands for ``game theory''}
\label{tab:Games}
\end{table}

\section{Methods}
\subsection{Training}

Each system was given 15 minutes to generate a hypothesis for each predicate. If the predicate was not learned the default hypothesis was \textit{true}. 

In \textbf{E1} The ILP systems were trained on 8 full game traces. They were each trained on intelligent random and mixed traces. The mixed traces were made up of a 50/50 mix of random and intelligent traces. 

In \textbf{E2} we trained each system on 8, 16 and 24 mixed traces where each set was made up of a 50/50 mix of random and intelligent traces. It was found that the results for 16 and 24 traces were very similar across all systems so no larger training sets were tested.

In the IGGP paper \cite{Cropper/IGGP} the IGGP task was defined across four predicates for each game: \textit{goal}, \textit{next}, \textit{terminal} and \textit{legal}. After testing the systems on all four predicates it was found that the score for the terminal predicates do not accurately represent the ability of each system. Generally the randomly generated game traces did not complete the game before the move limit was reached. Since the terminal predicate is true on the last game state in each game and the game state includes a move counter this allowed the systems to simply learn the correlation between the maximum move and the terminal predicate. They would learn the program: \verb|terminal :- true_step(MAX)| where \verb|MAX| is the maximum number of moves for the game. Since when a game is played randomly the max move count is almost always hit the predicate is often perfectly solved. However if when Sancho plays the game it solves it before the maximum move count is hit it is less likely to induce a correct rule. For this reason the \textit{terminal} predicate was not included in training data.

\subsection{Testing}

Each system was tested on 4 randomly generated and 4 intelligently generated traces. To evaluate the performance of the ILP systems on the training dataset we use two metrics: balanced accuracy and perfectly solved.
\\

\textbf{Balanced accuracy} In the datasets used for testing the ILP systems the vast majority of examples are negative. Balanced accuracy takes this into account when evaluating approaches. The generated logical hypothesis $H$ along with the background knowledge $B$ for the relevant game is tested. The test data is the set of combined positive and negative testing examples $E^+ \cup E^-$. We define the number of positive examples $p = |E^+|$, the number of negative examples $n = |E^-|$, the number of correctly predicted positives as $tp = |\{e\in E^+|B\cup H \models e\}|$ and the number of correctly predicted negatives as $tn = |\{e\in E^-|B\cup H \not\models e\}|$. The balanced accuracy is subsequently defined as $ba = 100 \cdot (tp/p + tn/n)/2$.

\textbf{Perfectly solved} This metric considers the number of predicates for which the ILP system correctly classified all examples. It is equivalent to the number of games with a balanced accuracy score of 100. This metric is important since for all predicates there exists an exact solution (the rules that were used to generate the examples). A system that has correctly predicated 99\% of the results is not nearly as useful as one that predicts 100\%.
\\

In the results section (chapter \ref{ch:results}) we present only the aggregate scores in the results since the full results are too large for this paper. To evaluate the systems we compare the average balanced accuracy and the number of perfectly solved games when tested on optimal and random traces.

\chapter{Results}\label{ch:results}
We now describe the results of testing the ILP systems. We conducted the experiments in accordance with the methods described in chapter \ref{ch:methodology}.

To recap the research questions are as follows:
\begin{itemize}
	\item \textbf{Q1} - Does varying the quality of game traces influence the ability for learners to solve the IGGP problems? Specifically, does the quality of game play affect predictive accuracy?
	\item \textbf{Q2} - Does varying the amount of game traces influence the ability for learners to solve the IGGP problems? Specifically, does the quality of game play affect predictive accuracy?
	\item \textbf{Q3} - Can we improve the performance of a learner by mixing the quality of traces?
\end{itemize}

Experiment \textbf{E1} answers questions Q1 and Q3 and experiment \textbf{E2} answers question Q2.

\section{Results summary}

\subsection{E1: Varying the quality of game traces}
In experiment E1 we train and test each system on random and intelligent traces as well as a 50/50 mixture of both. Table \ref{tab:E1-BA} shows the average balanced accuracy of each test of the systems. Table \ref{tab:E1-P} shows the number of games for which the rule was perfectly learned for each test. The the differences for each system in the average balanced accuracy for the different training/testing combinations were unpronounced. To show this we use the $\chi^2$ (chi squared) test to calculate the statistical significance of the results. The biggest difference between any two balanced accuracy testing results between two distributions for a single system was for Aleph where it was trained on Sancho traces and then tested on Random and Sancho traces. The $\chi^2$ test was performed on these two scores. The result was a p-value of 0.6723. This value is well above any reasonable threshold for statistical significance meaning the results can be viewed as not significant. Despite this there are still differences worth discussing between the different results.

\begin{table}[]
	\begin{tabular}{|c|c|c|c|c|c|c|}
		\hline
		\multicolumn{7}{|c|}{Balanced Accuracy}                                                                           \\ \hline
		\multirow{2}{*}{Systems} & \multicolumn{2}{c|}{Random} & \multicolumn{2}{c|}{Sancho} & \multicolumn{2}{c|}{Mixed} \\ \cline{2-7}
		& Random       & Sancho       & Random       & Sancho       & Random       & Sancho      \\ \hline
		Metagol                  & 51           & 51           & 51           & 51           & 51           & 51          \\ \hline
		Aleph                    & 66           & 65           & 65           & 72           & 67           & 66          \\ \hline
		ILASP                    & 72           & 72           & 71           & 72           & 73           & 73          \\ \hline
	\end{tabular}
	\caption{The average balanced accuracies across all games and predicates for each system in E1. Each row gives the balanced accuracies of one system. Each was trained on random, Sancho generated and mixed traces then tested against random and Sancho generated. The top of the two header rows gives the training distribution and the lower gives the test distribution}
	\label{tab:E1-BA}
\end{table}

\begin{table}[]
	\begin{tabular}{|c|c|c|c|c|c|c|}
		\hline
		\multicolumn{7}{|c|}{Perfectly Solved}                                                                            \\ \hline
		\multirow{2}{*}{Systems} & \multicolumn{2}{c|}{Random} & \multicolumn{2}{c|}{Sancho} & \multicolumn{2}{c|}{Mixed} \\ \cline{2-7}
		& Random       & Sancho       & Random       & Sancho       & Random       & Sancho      \\ \hline
		Metagol                  & 0            & 0            & 0            & 0            & 0            & 0           \\ \hline
		Aleph                    & 1            & 0            & 0            & 4            & 1            & 0           \\ \hline
		ILASP                    & 7            & 8            & 5            & 10           & 5            & 8           \\ \hline
	\end{tabular}
	\caption{The number of perfectly solved scores for each system trained and tested as described in E1}
	\label{tab:E1-P}
\end{table}

\subsubsection{Aleph}
Throughout all experiments conducted the programs learned by Aleph consisted almost entirely of a disjunction of cases seen in testing. It appears that the most general clause (see section \ref{sec:aleph}) that could be found for a lot of the examples was also the most specific clause. This simplistic method resulted in programs that performed better when the training and testing data was closely matched. This can be seen in the results for Aleph when trained and tested on Sancho generated traces. For some games all Sancho generated traces were the same, in \textit{fizz buzz} Sancho always says the correct thing and in the \textit{prisoner dilemma} both Sancho players always defect. This allowed Aleph to achieve a perfect score on the \textit{next} predicate. In cases where the training data is diverse Aleph learns a program that over-approximates the solution doing well on the tests with positive examples and badly on the tests with negative examples.

Generally Aleph was best at learning the \textit{goal} predicate. Unlike \textit{next} which takes a game state and player moves as input the \textit{goal} predicate often only takes the player role. The limited number of output values, 0 to 100, also meant the output space was smaller than the other predicates. The program Aleph produced would often simply associate each role to all the different goal values it had at any point in the training data. This is a subsection of the goal values it learned for \textit{checkers}:
\begin{verbatim}
goal(red,0).
goal(black,0).
goal(black,8).
goal(red,8).
\end{verbatim}
This method does not capture the original game rule at all but achieves reasonable scores in testing.

Aleph did not do particularly better when trained on the mixed training set. It achieved similar scores to the other tests apart from when it was trained and tested on intelligent traces.

\subsubsection{Metagol}
A limitation with the Metagol system is that if it cannot learn a rule that covers 100\% of the positive training examples and 0\% of the testing examples it will not output any rule at all. With the time limitations placed on the experiments Metagol rarely managed this. If some rule contains a special case such as \textit{castling} in \textit{checkers} Metagol may have learned the general \textit{next} predicate but would not give any output since this case was not covered. The search space scales exponentially with the size of the rules to be learned.

Metagol never learned any \textit{legal} or \textit{goal} predicates. The \textit{goal} predicate is often relatively complex. It can rely on game concepts which would be obvious to humans but hard to learn for an ML system. For example the idea of three of the same symbol in a row, column or diagonal in \textit{tic tac toe}. There might not be enough training examples to guide Metagol to concepts such as these. The \textit{legal} predicate is usually even more complex than the goal, however in some cases the legal moves were constant across all states. It is possible that the metarules supplied to Metagol for this experiment were not ideal for this task.

Where Metagol learned the \textit{next} predicate it often only managed the simple parts of it. The predicate is split up into different parts such as \textit{next\_state} and \textit{next\_step} to reflect the different predicates that change each move. If one of the predicates was \textit{next\_step(N)} where \textit{N} increased by 1 each turn Metagol managed to learn that however it failed to produce any program that was more than two or three lines long.

\subsubsection{ILASP}
ILASP performed by far the best out of the three systems. The differences in results between the systems trained on Sancho and random data was not large. Out of them it scored the best when trained and tested on Sancho traces presumably for the same reasons that Aleph did (the testing data was very similar to training data). The margin by which this scored better however was almost negligible. There was generally less than 2\% difference between the balanced accuracies for each predicate of each game.

ILASP scored several percentage points better across the board when trained on the mixed data. ILASP benefited the most from the combination of training data out of all the systems. It scored better on the \textit{goal} and \textit{next} predicates but not on the \textit{legal} predicate. This makes sense because the legal predicate changes the least between the intelligent and random traces, the same complexity of rule still generally has to be learned in each. The \textit{next} predicate for \textit{tic tac toe} was learned a lot better by ILASP on the mixed data. It is unclear exactly why this is the case but it does appearer ILASP learns better with greater diversity of examples.

\subsection{E2: Varying the amount of game traces}
In experiment E2 each system was trained three times on mixed traces. The first time with a total of 8 traces, the second with 16 traces and the third with 24 traces. When designing this experiment it was assumed that the more traces they were given to train on the better the systems would perform. However, as can been see in the scatter graph of the results this was not the case.
\\\\\\
\begin{tikzpicture}
\begin{axis}[scatter/classes={
	a1={mark=square*,blue},%
	a2={mark=triangle*,blue},%
	b1={mark=square*,red},%
	b2={mark=triangle*,red},%
	c1={mark=square*,green},%
	c2={mark=triangle*,green}},
legend style={at={(1.35,0.7)},
	anchor=north},
xtick=data,
ylabel={Balanced Accuracy},
xlabel={Number of traces}]
\addplot[scatter,only marks,
scatter src=explicit symbolic] coordinates {
	(8,51)     	[a1]
	(8,51)     	[a2]
	(16,51)    	[a1]
	(16,51)    	[a2]
	(24,51)    	[a1]
	(24,51)		[a2]
	(8,67)		[b1]
	(8,66)		[b2]
	(16,70)		[b1]
	(16,65)		[b2]
	(24,70)		[b1]
	(24,65)		[b2]
	(8,73)		[c1]
	(8,73)		[c2]
	(16,50)		[c1]
	(16,50)		[c2]
	(24,50)		[c1]
	(24,50)		[c2]
};
\legend{\ \ \ \ \ Metagol tested on Random,\ \ \ Metagol tested on Sancho,\ \ Aleph tested on Random,Aleph tested on Sancho, \ \ \ ILASP tested on Random,\ ILASP tested on Sancho,d}
\end{axis}
\end{tikzpicture}
\\\\
When ILASP and Metagol have not been given enough time to fully learn a predicate they output nothing. Both systems scale exponentially in time taken according to the size of the input. Whilst for Metagol it didn't have time even for the 8 traces ILASP ran out of time only on the 16 and 24 traces training sets. This had the effect for ILASP of a large drop in effectiveness. For 16 and 24 the default program consisting of just \textit{true} was used. This gave the balanced accuracy of 50 for each predicate. Metagol still managed to learn the simple predicates that it had managed on the 8 trace training set however did not increase at all in effectiveness over the rest of the experiment.

Unlike the other systems Aleph was consistently able to process the entire input since its method of learning is less computationally intense. When tested on the random traces Aleph did better with the increased number of traces. This was most likely due to the fact that more cases were encountered in the training data. Since Aleph usually ends up putting the cases it encounters straight into the program this simply meant there was a greater number of situations in the testing set that Aleph had seen before and was thus correctly able to classify. When tested on the intelligent traces Alephs score was lower. This is possibly due to the maximum size of Aleph programs taking effect. The clauses relevant to the intelligent traces may be pushed out of the program by the vast number of clauses taken from the random gameplay examples.

If this experiment were to be conducted again much more enlightening results than these could be obtained by increasing the time given to each system to learn each predicate or even using a system with greater computational power to run the experiments.

\chapter{Conclusions}
In this paper we have compared the ability of ILP systems to learn the rules of a game from gameplay observations in the IGGP framework. Three ILP systems have been trained on a range of different datasets. These datasets consisted of random  gameplay, intelligent gameplay as generated by the Sancho system, and a mix of both. The generated hypotheses of the learners have been tested on random and intelligent gameplay from the same distributions as the training data. Differences in the effectiveness of the learned programs in correctly classifying legal and illegal gameplay were observed however there was not enough statistical significance in the results to disprove the null hypothesis.

To test the significance of the size of the training dataset on the ability of the ILP systems to learn the systems were trained on varying numbers of game traces. The results of the experiment were inconclusive as not all the systems had enough time to learn a hypothesis when trained on the large datasets.

\subsubsection{Limitations}
There are several aspects that could be improved on in this paper.
\\

\textbf{Computational resources}
The computational resources and thus the time given for each system to learn the rules were severely limited for the experiments in this paper. If they were to be conducted again with greater computational resource with more time given to each system more significant results may be achieved.

\textbf{ILP systems} In this paper we only test on three ILP systems, it is clear a more representative result could be obtained by testing on more systems. Techniques such as probabilistic \cite{Bellodi/Probablistic,Raedt/Probabalistic} and interactive \cite{Raedt/Interactive} learning are not used by any of the systems tested here. ILASP is the only system designed to handle noisy data \cite{MarkLaw/ILASP2i}, it would be interesting to try others with this approach \cite{Oblak/Noise,Evans/Noise}. The systems used also have many customisable settings for example Metagols metarules which have major implications for the programs learnt \cite{Cropper/Metarules}. Aleph has many different search methods on the hypothesis space many of which may yield better results than the default algorithm used in this paper.

\textbf{Generating traces} Whilst Sancho generates good examples of intelligent play it would be interesting to see traces generated by other general game playing methods \cite{Park/GGPAdvances,Kowalski/GGP}. For some games there exists a provably optimal set of moves in some cases such as for \textit{eight puzzle} Sancho generates these moves (see section \ref{sec:sancho}) however it does not for all such games \cite{Schaeffer/Checkers}. Future research could compare the effects on the learned hypotheses when trained on optimal data. It may also be more insightful to look at the difference between human generated data and random or optimal.

\textbf{Other machine learning systems} There exists a huge variety of machine learning systems other than ILP. It is possible some of these may exhibit more of a bias toward one of the training sets. Testing a neural network or genetic algorithm based approach may provide insightful results.

\textbf{Intelligent versus random in other contexts} There are plenty of other domains other than game playing in which datasets can be considered intelligent or random. An example of this might be training a car to drive where you could train it on different quality levels of driving or training a facial recognition system on poor or high quality photos. These would prove an interesting area to study.

\textbf{Levels of intelligence} So far we have only considered a single level of intelligent play. It is clear that in humans for any given game we do not simply have good players and random players. There is a scale of skill. The ELO system captures this concept well and is used as a generic measure of skill for zero sum games but has been extended to more \cite{ELO}. It would be informative to pick a game and use game traces generated by player of a range different ELO score. It would then be clear if the systems in fact learned best from a particular level of play rather than just optimal or random. We expect that the best level of play for training would differ among games as has been shown in this paper. However it would still be useful to discover what level was optimal for any game. 

\textbf{How do humans compare} We have spent a lot of this paper investigating whether machines learn better on intelligent gameplay or random gameplay with the assumption that a human would learn better from watching another human play than random legal moves. Whilst in reality we might generally choose to watch a human playing when trying to learn a game rather than observing random legal moves it may not be the most effective way. Any real world example of this is almost always accompanied by additional information. For example humans would often have access to some description of the rules or someone explaining the edge cases that may not occur. It is hard to compare how humans learn to how machines learn since the background knowledge of a human is so much greater. It would be interesting to see how humans would fare in similar experiments to those in this paper. Looking into this question would give new insights into the results of this paper. As with the similar question posed in this paper the experiment does not have clear predicted outcome.

\bibliography{ms}
\bibliographystyle{plain}
\end{document}